\documentclass{article} 
\usepackage{iclr2017_workshop,times}
\usepackage{hyperref}
\usepackage{url}
\usepackage{graphicx}
\usepackage{amsmath}
\usepackage{multirow}
\usepackage{grffile}

\usepackage{amsmath}
\usepackage{amssymb}
\usepackage{amsthm}

\newcommand{\myvec}[1]{\mathbf{#1}}

\newcommand{\vd}{\myvec{d}}

\newcommand{\vx}{\myvec{x}}

\newcommand{\unif}{\mathrm{Unif}}

\newcommand{\hmm}[1]{\ifnum\ifhmode\spacefactor\else2000\fi>1000 \uppercase{#1}\else#1\fi}

\DeclareMathOperator*{\argmax}{arg\,max}
\DeclareMathOperator{\sign}{sgn}

\title{Delving Into Adversarial Attacks on Deep Policies}

\author{Jernej Kos \\
National University of Singapore\\
\And
Dawn Song \\
University of California, Berkeley\\
}

\begin{document}

\maketitle

\begin{abstract}
Adversarial examples have been shown to exist for a variety of deep learning architectures.
Deep reinforcement learning has shown promising results on training agent policies directly on raw inputs such as image pixels.
In this paper we present a novel study into adversarial attacks on deep reinforcement learning polices.
We compare the effectiveness of the attacks using adversarial examples vs. random noise. We present a novel method for reducing the number of times adversarial examples need to be injected for a successful attack, based on the value function. We further explore how re-training on random noise and FGSM perturbations affects the resilience against adversarial examples.
\end{abstract}

\section{Introduction}

Adversarial examples have been shown to exist for a variety of deep learning architectures.
They are small perturbations of the original inputs, often barely visible to a human observer, but carefully crafted to misguide the neural network into producing incorrect outputs.
Seminal work by~\citet{szegedy2013intriguing} and ~\citet{goodfellow2014explaining}, as well as much recent work, has shown that adversarial examples are abundant and finding them is easy.
Deep neural networks have been used in deep reinforcement learning (DRL) with promising results on training policies directly on raw inputs such as image pixels.
One of the most successful algorithms for training deep policies is A3C~\citep{mnih2016asynchronous}, which enables asynchronous updates of policy weights, leading to an efficient parallel implementation.
As the policies may drive various autonomous agents such as self-driving cars in the real world, adversarial attacks may be of even greater importance.

Our paper is among the first to investigate adversarial examples on DRL policies, showing that these deep policies are easily fooled by adversarial attacks with very small adversarial perturbations. 
In addition, in this paper, we examine three new dimensions about adversarial attacks on DRL policies that no other work has not addressed before.
First, we compare adversarial examples to random noise and show that the former are an order of magnitude more effective for attacking DRL policies.

Another important dimension with DRL systems is time.
If the attacker needs to inject adversarial perturbations less frequently, then the attack is easier to perform.
To this end, we explore using the policy's value function as a guide for when to inject perturbations.
Our experiments show that with guided injection, the attacker can succeed with injecting perturbations in only a fraction of the frames, and is more effective than injecting perturbations with similar frequency but without the guidance.
Our results show that adversarial attacks can be much more complex in the reinforcement learning setting than other settings previously studied such as image classification.

The third dimension is policy resilience through re-training.
We present preliminary results showing that the agents are able to become more resilient to fast-gradient sign method (FGSM) attack under re-training with both random noise and FGSM perturbations, while re-training with FGSM perturbations may be more effective than re-training with random noise.
The re-trained agent may still be vulnerable to other attack methods such as optimization-based attacks, however, these other attack methods are much slower to perform, often rendering the attacks extremely slow especially for the agent setting.

Concurrently and independently from our work (submission to the same ICLR workshop), \citet{huang2017adversarial} also presented a study into adversarial attacks on DRL policies, showing when an attacker injects small adversarial perturbations into every frame, the learned agent will fail.

Due to space limit, we focus on agents trained on the Atari Pong task using the A3C algorithm and FGSM adversarial perturbations.
Our work is a first step towards better understanding of the challenges and limitations of DRL under adversarial inputs.  

\section{Study Objectives}

\paragraph{Attack Effectiveness of Adversarial Examples vs. Random Noise}

We study how injecting random noise into the environment compares to injecting FGSM adversarial perturbations.

\paragraph{Using the Value Function to Guide Adversarial Perturbation Injection}

We want to see if reducing the frequency of adversarial perturbation injection can still generate an effective attack.
We study three different methods: a) we only inject an adversarial perturbation every $N$ frames and the intermediate frames are without any perturbation, b) we only recompute an adversarial perturbation every $N$ frames and inject the last computed perturbation in the intermediate frames; and c) we use the value function, computed over the original input, in order to estimate when to inject the adversarial perturbation for it to be most effective, and only inject the adversarial perturbation when this estimate is above a certain threshold.

\paragraph{Effectiveness of Re-training with Adversarial Examples and Random Noise}

We study whether the agents can be re-trained on an environment with injected random noise or adversarial perturbations in order to make them more resilient against further adversarial perturbations.
Additionally, we study whether this obtained resilience transfers to environments with different magnitudes and different types of perturbations (e.g., is an agent trained on random noise any more resilient to FGSM adversarial perturbations).

\section{Experimental Evaluation}

To perform our experiments, we use a TensorFlow implementation of the A3C~\citep{mnih2016asynchronous} algorithm.
We evaluate the method on the Atari Pong task, where the initial input image pixels are cropped and scaled to 42x42. Finally, luminosity is computed from RGB values, giving us frame dimensions of 42x42x1.

To generate adversarial perturbations, we use the fast gradient sign method (FGSM) initially developed by~\citet{goodfellow2014explaining}.
FGSM requires a loss function $J(\theta, x, y)$ in order to compute its gradient $\nabla_x$.
We use the cross-entropy loss between $y$ (a vector of logits, representing weights for each action, produced by the policy) and the one-hot encoding of $\argmax y$.
This means that the attack attempts to generate an input which moves the policy output away from the optimal action.

In all our experiments, the agent is first trained on a baseline (non-noisy) environment until it achieves an optimal reward for a number of episodes (baseline agent).
Then, for generating the FGSM perturbations we set an appropriate $\epsilon$ and compute $\epsilon \sign \nabla_x J(\theta, x, y)$.
For generating random noise, we sample from a uniform distribution $\unif(0, \beta)$, where we set $\beta$ based on the required intensity.

\paragraph{Attack Effectiveness of Adversarial Examples vs. Random Noise}

The baseline agent is evaluated on a modified version of the environment, where either random noise or FGSM perturbation is injected on every frame.
Figure~\ref{fig:fixed-random-vs-adversarial} in Appendix shows the difference in attack effectiveness between random noise and FGSM perturbations.
While low levels of random noise ($\beta \leq 0.02$) do not impact the agent's performance much, using random noise of greater magnitude ($\beta \geq 0.05$) severely degrades performance.
FGSM adversarial perturbations are orders of magnitude more effective than random noise for successful attacks, succeeding on attacking the baseline agent at much lower perturbation levels.

\paragraph{Using the Value Function to Guide Adversarial Perturbation Injection}

\begin{figure}
\begin{center}
\includegraphics[scale=0.22]{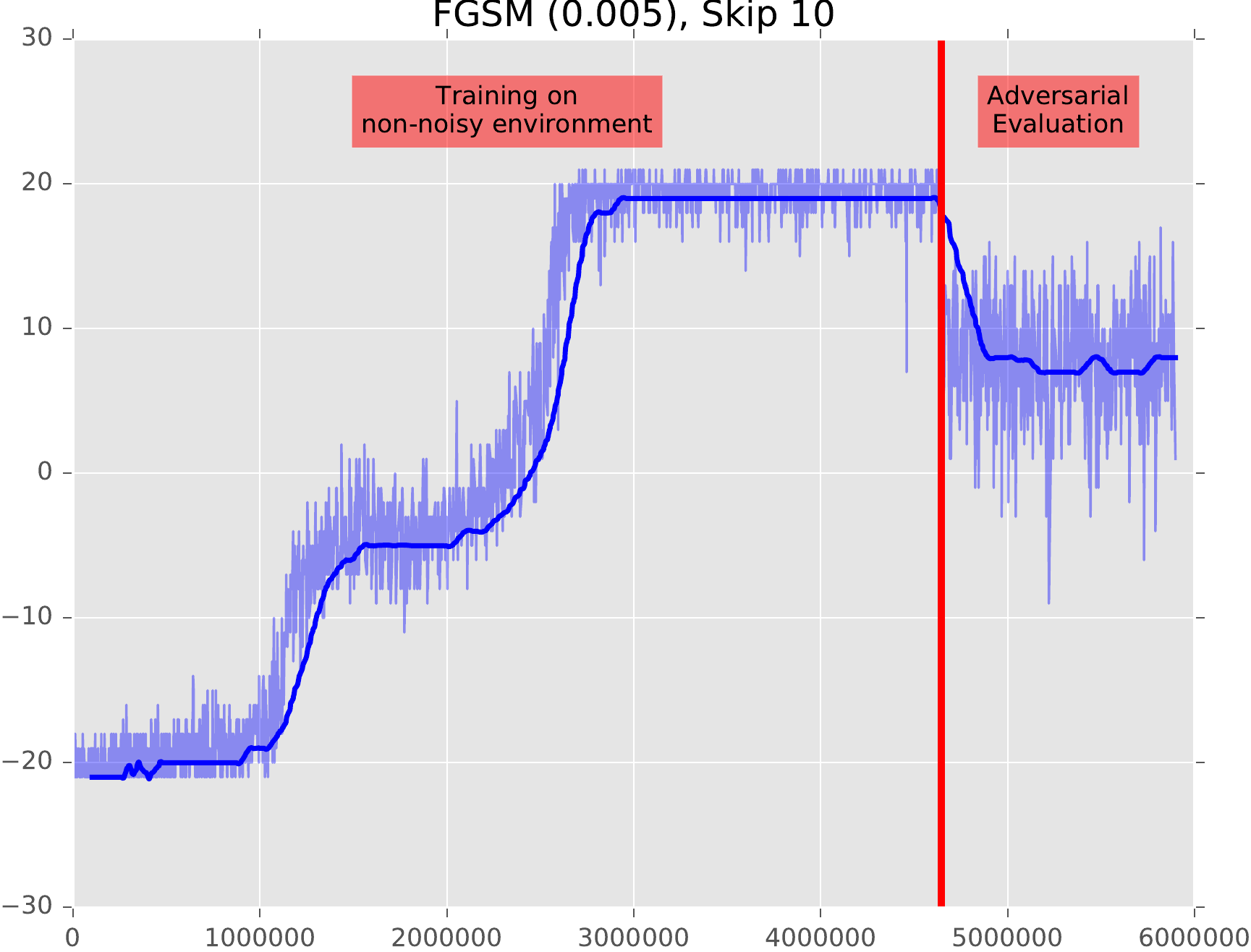}
\includegraphics[scale=0.22]{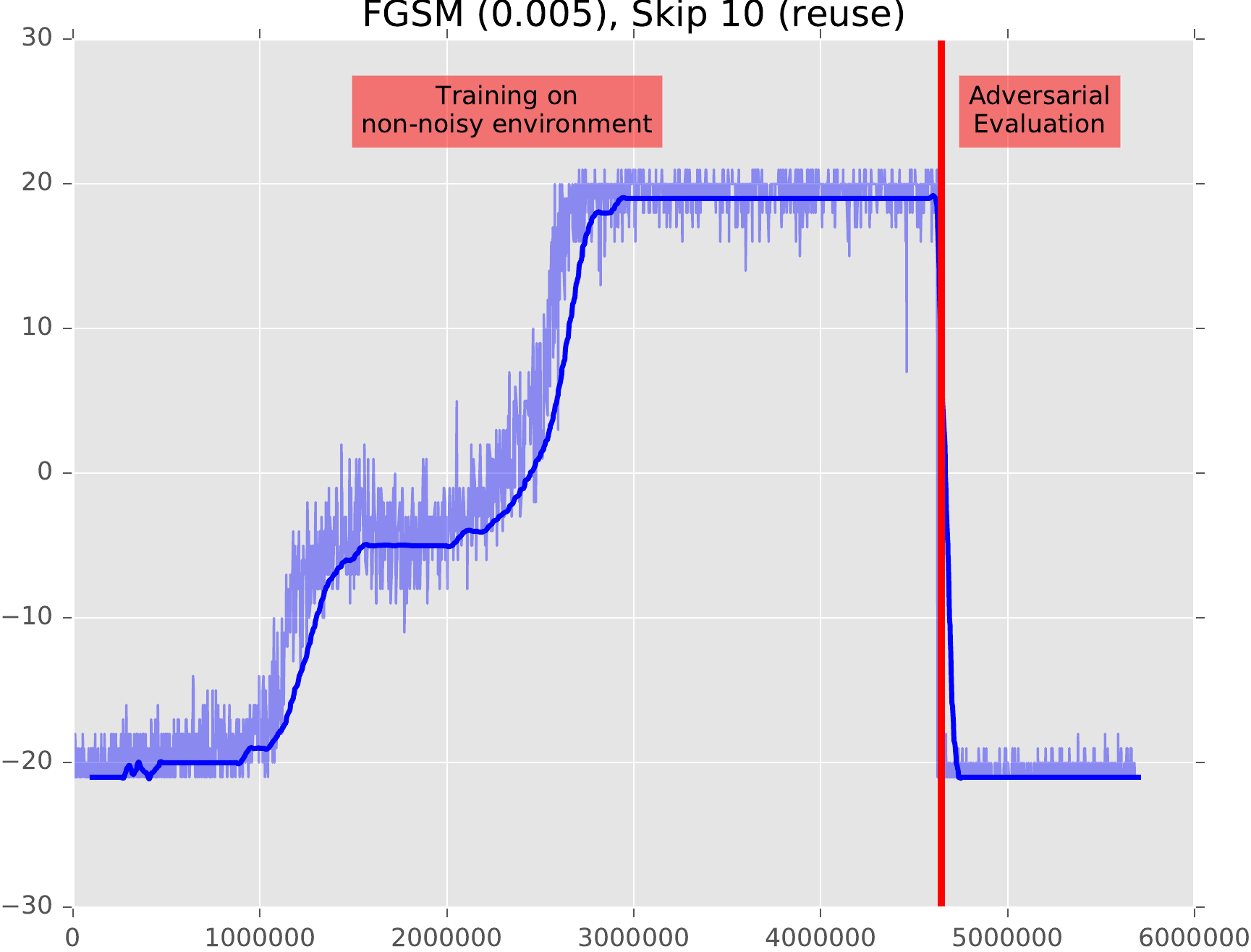}
\end{center}
\caption{Attack effectiveness when FGSM perturbations are only injected every 10th frame (left) and when the perturbations are only recomputed every 10th frame, but reused in the intermediate frames (right).}
\label{fig:reuse-skip-attack}
\end{figure}

First, we explore how the frequency of injecting adversarial perturbation affects attack success. In this experiment, we either inject FGSM perturbations only every tenth frame and use original frames in-between, or recompute perturbations every tenth frame and use the last computed perturbation in-between.
All experiments were performed with $\epsilon$ set to $0.001$.
Our results hwo that only injecting FGSM perturbations on every tenth frame does not seem to be a particularly effective attack (Figure~\ref{fig:reuse-skip-attack}, left).
On the other hand,  recomputing perturbations every tenth frame and reusing the previous perturbation in intermediate frames is equally effective as the original attack (Figure~\ref{fig:reuse-skip-attack}, right).

\begin{figure}[h]
\begin{center}
\includegraphics[scale=0.22]{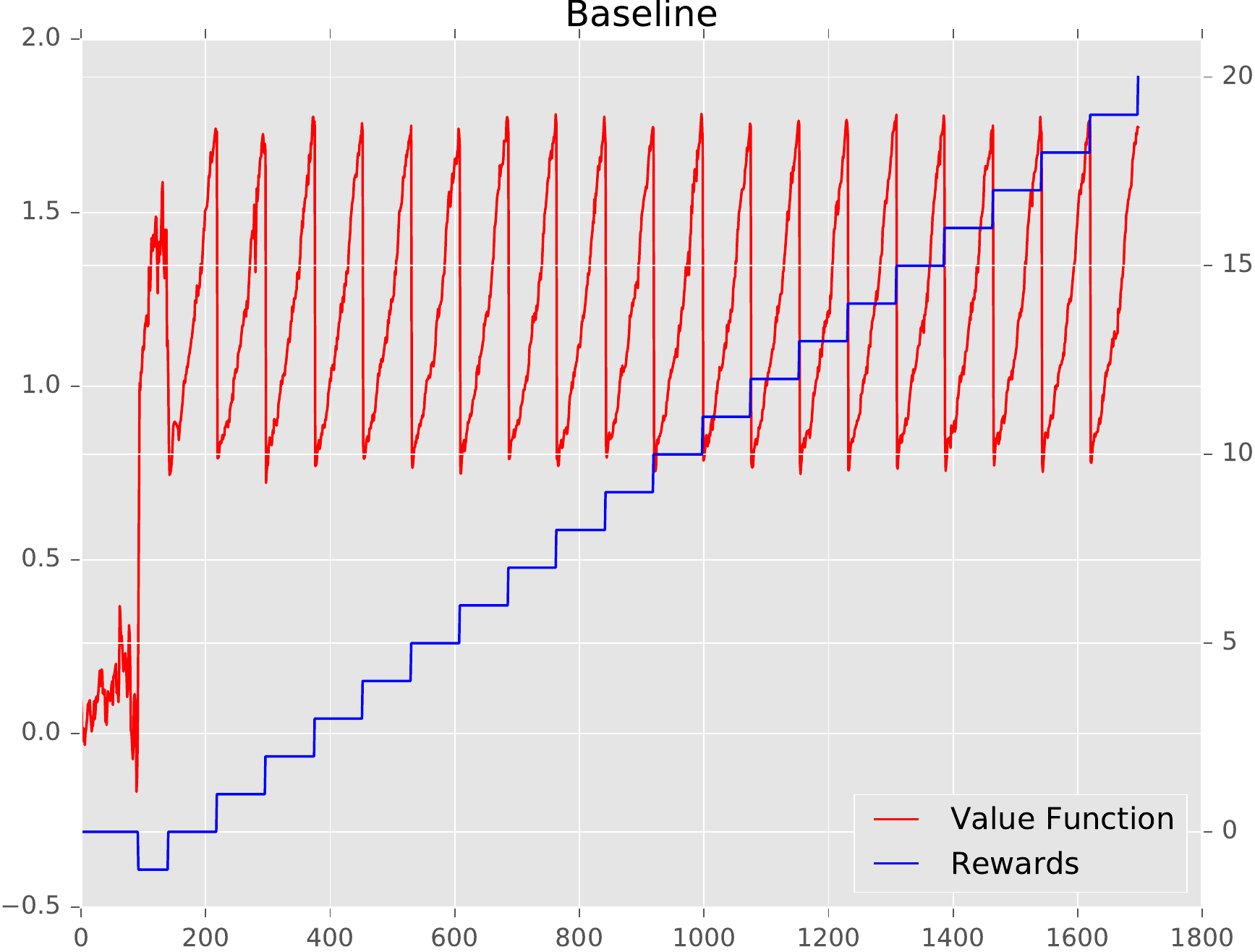}
\includegraphics[scale=0.22 ]{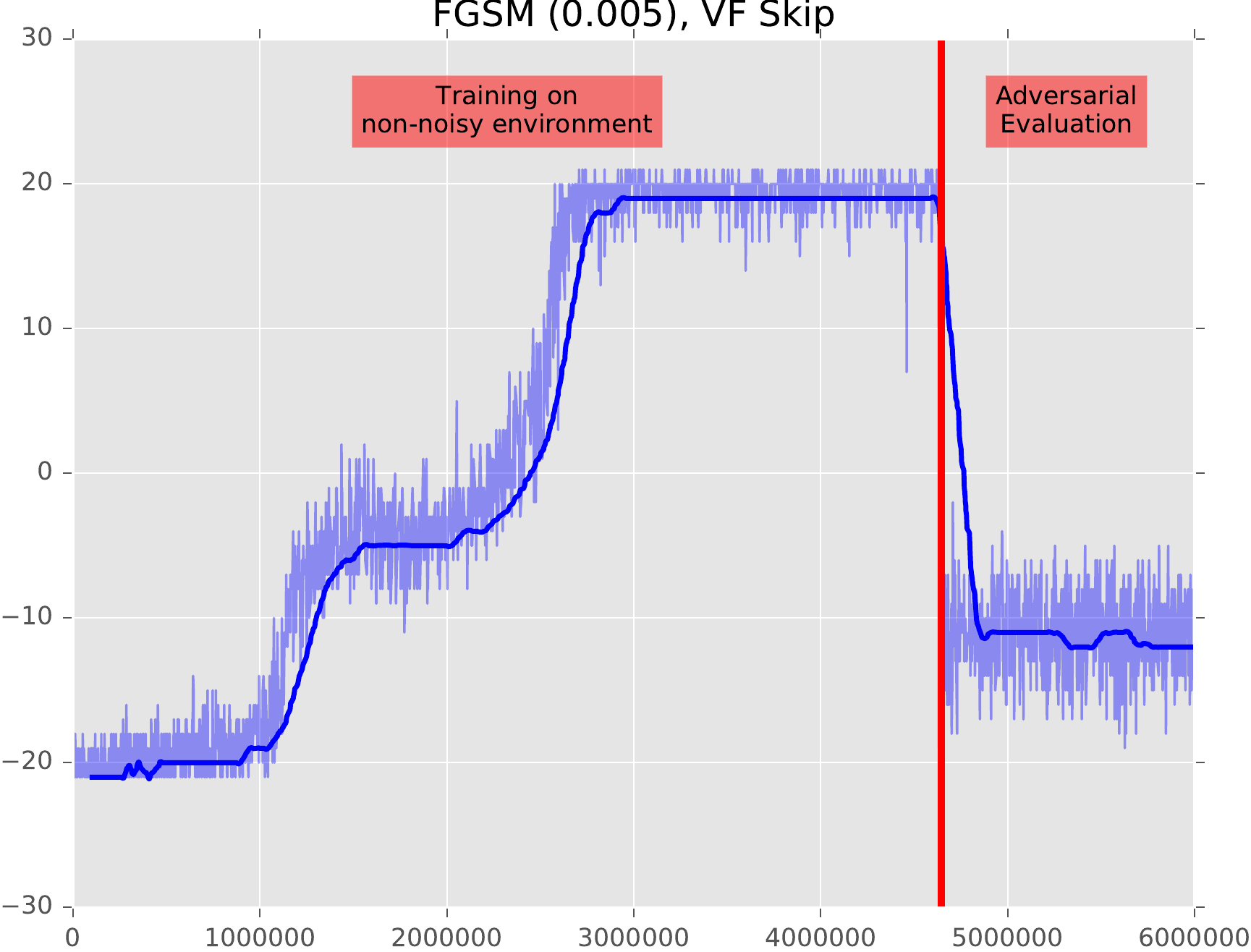}
\end{center}
\caption{Policy's value function approximation during one baseline episode (left).
Effectiveness of injecting FGSM perturbations only in frames where the value function is above a threshold (right).}
\label{fig:vf-attack}
\end{figure}

We also develop an attack method (VF) where we inject adversarial perturbations only when the value function, computed over the original frame, is above a certain threshold (in this experiment we set the threshold to $1.4$).
The reasoning behind this is that we only want to disrupt the agent in crucial moments, when it is close to achieving a reward.
Figure~\ref{fig:vf-attack} shows the effectiveness of this method, demonstrating that the VF attack method is very effective while only injecting adversarial perturbations in a fraction of the frames.
We can compare the VF method against blindly injecting perturbations on every tenth frame (Figure~\ref{fig:reuse-skip-attack}, left).
Even though both methods inject perturbations a similar number of times on average during one episode ($120$ for the VF method and $125$ for the blind method), the VF method shows to be much more effective. This demonstrates that an attacker can use the value function to conduct a more efficient attack than the traditional attack method where the adversarial perturbation is injected in every frame (as in \citep{huang2017adversarial}). This also shows that adversarial attacks can be much more complex in the reinforcement learning setting than other setting previously studied such as image classification.

\paragraph{Effectiveness of Re-training with Adversarial Examples and Random Noise}

Finally, we also explore if the agents can be re-trained to improve resilience to both random noise and FGSM adversarial perturbations.
We additionally explore if this resilience then transfers to different magnitudes and types of perturbations.
During these experiments, after the initial training in non-noisy environment, the agent is first allowed to re-train while we inject random noise or FGSM perturbations on each frame.
After the agent achieves good performance, it is then frozen and evaluated in a new noisy environment, either with random noise or FGSM perturbations.

Figure~\ref{fig:training-random-vs-adversarial} in Appendix shows that in this setting, the baseline agent can be resilient against certain levels of FGSM perturbations after re-training on a noisy environment for a number of episodes, with sufficient level of random noise or FGSM perturbations added during re-training.
More interestingly, our experiment shows that the re-trained agent is also resilient against FGSM perturbation of much greater (or smaller) magnitude than the magnitude of the FGSM perturbations were used during re-training.
We also visualize the actions predicted by the policy in image space (see Figure~\ref{fig:action-boundary}).

\bibliography{extended_abstract}
\bibliographystyle{iclr2017_workshop}

\appendix
\section{Appendix}

\begin{figure}[h]
\begin{center}
\includegraphics[scale=0.30]{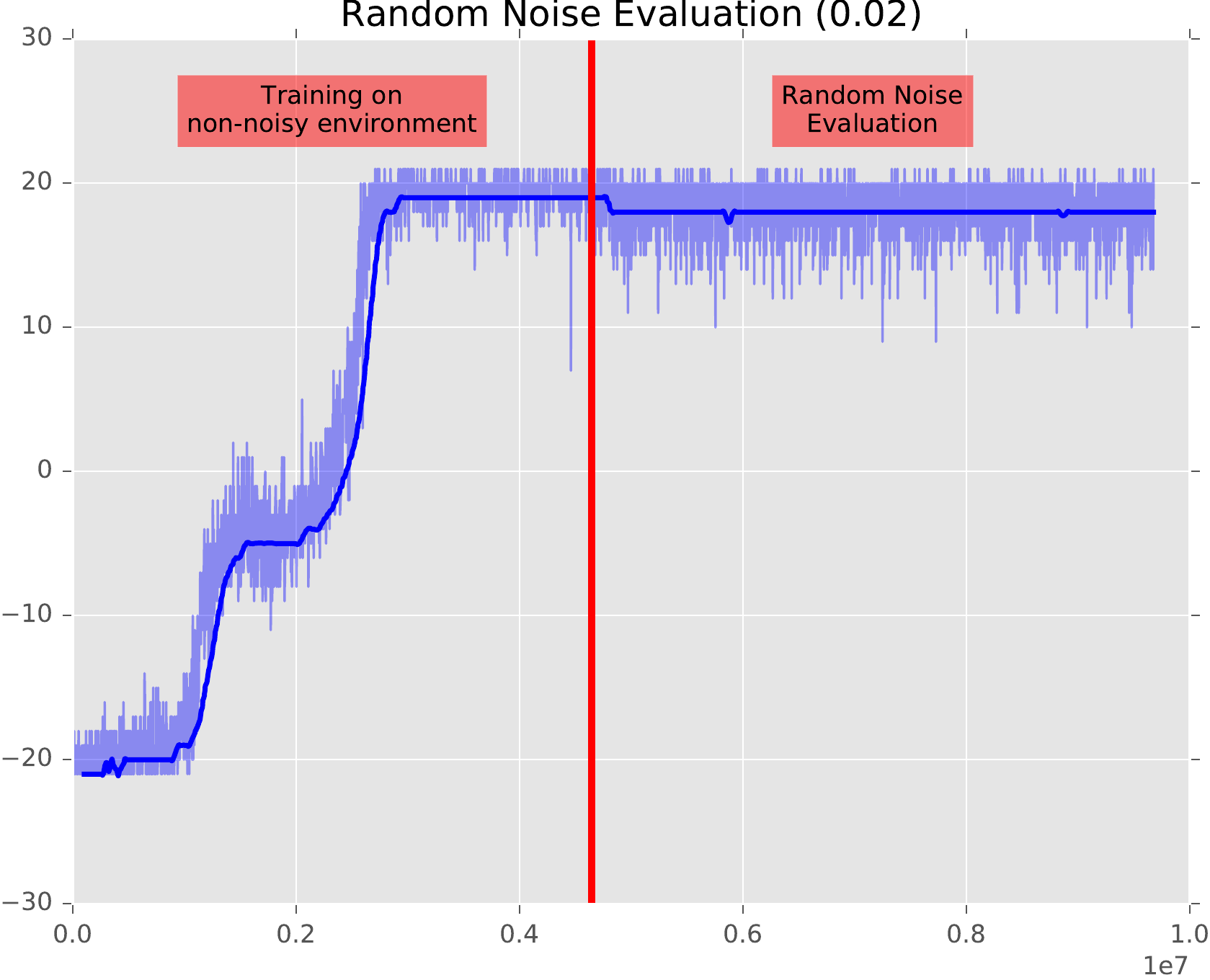}
\includegraphics[scale=0.30]{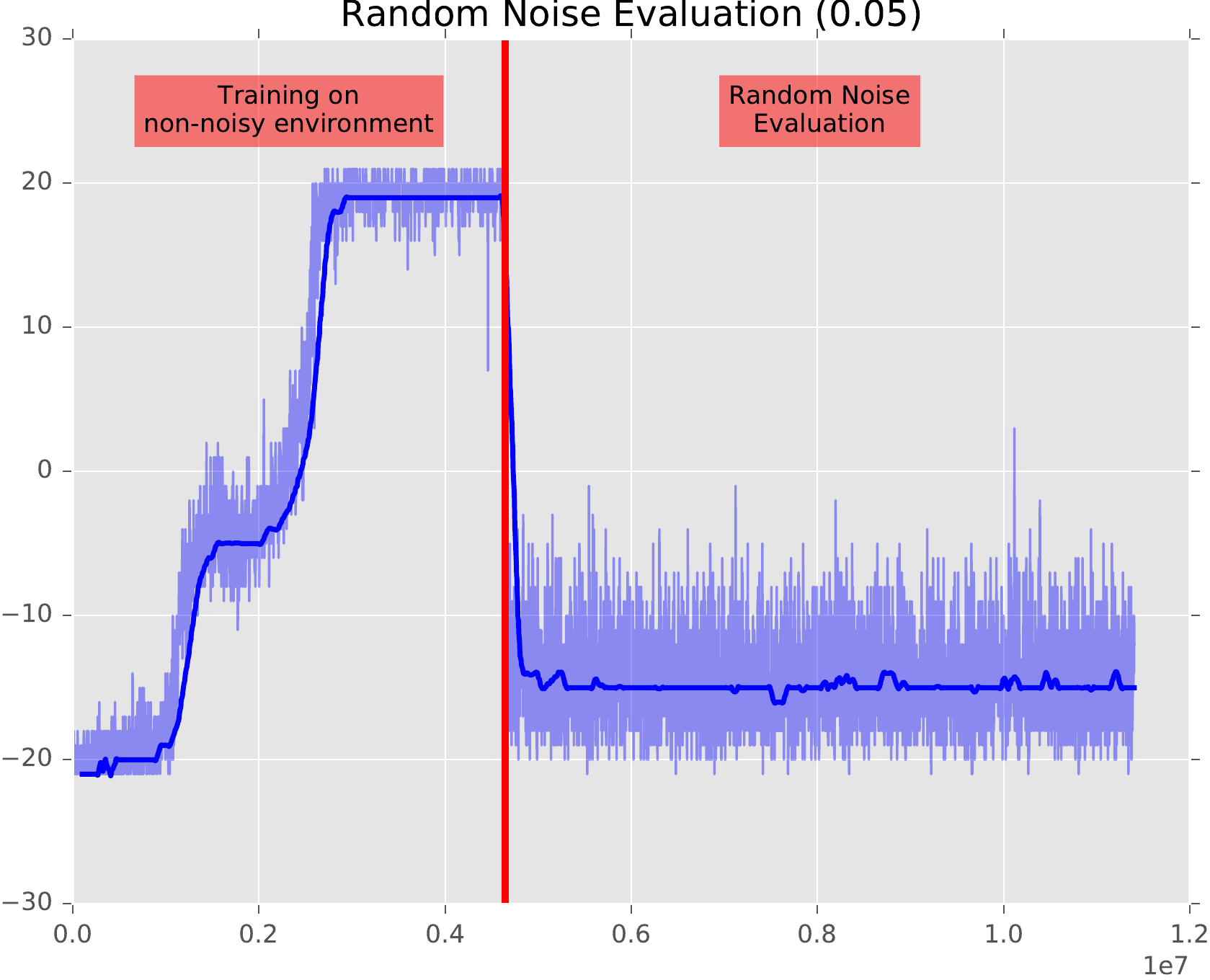} \\
\includegraphics[scale=0.30]{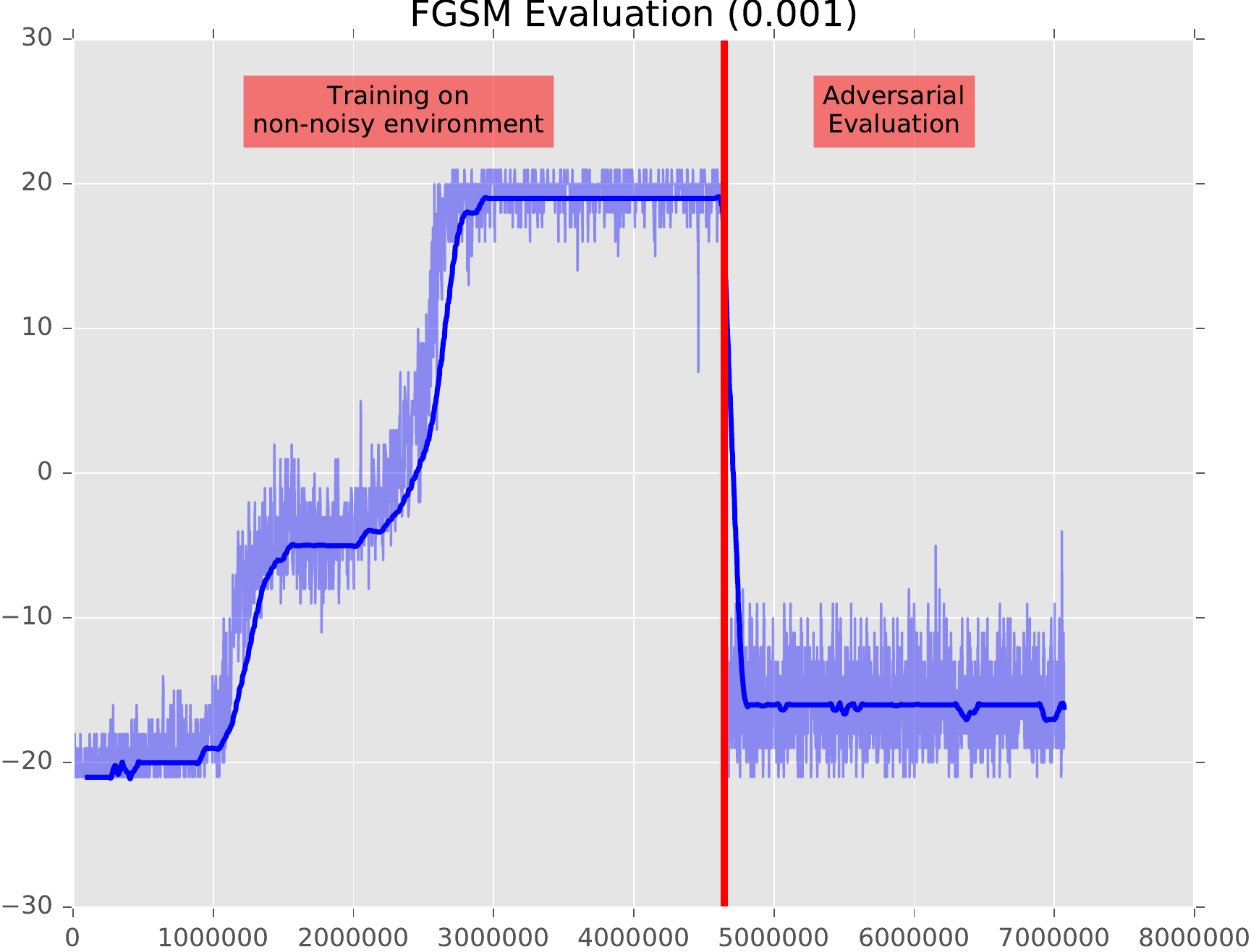}
\includegraphics[scale=0.30]{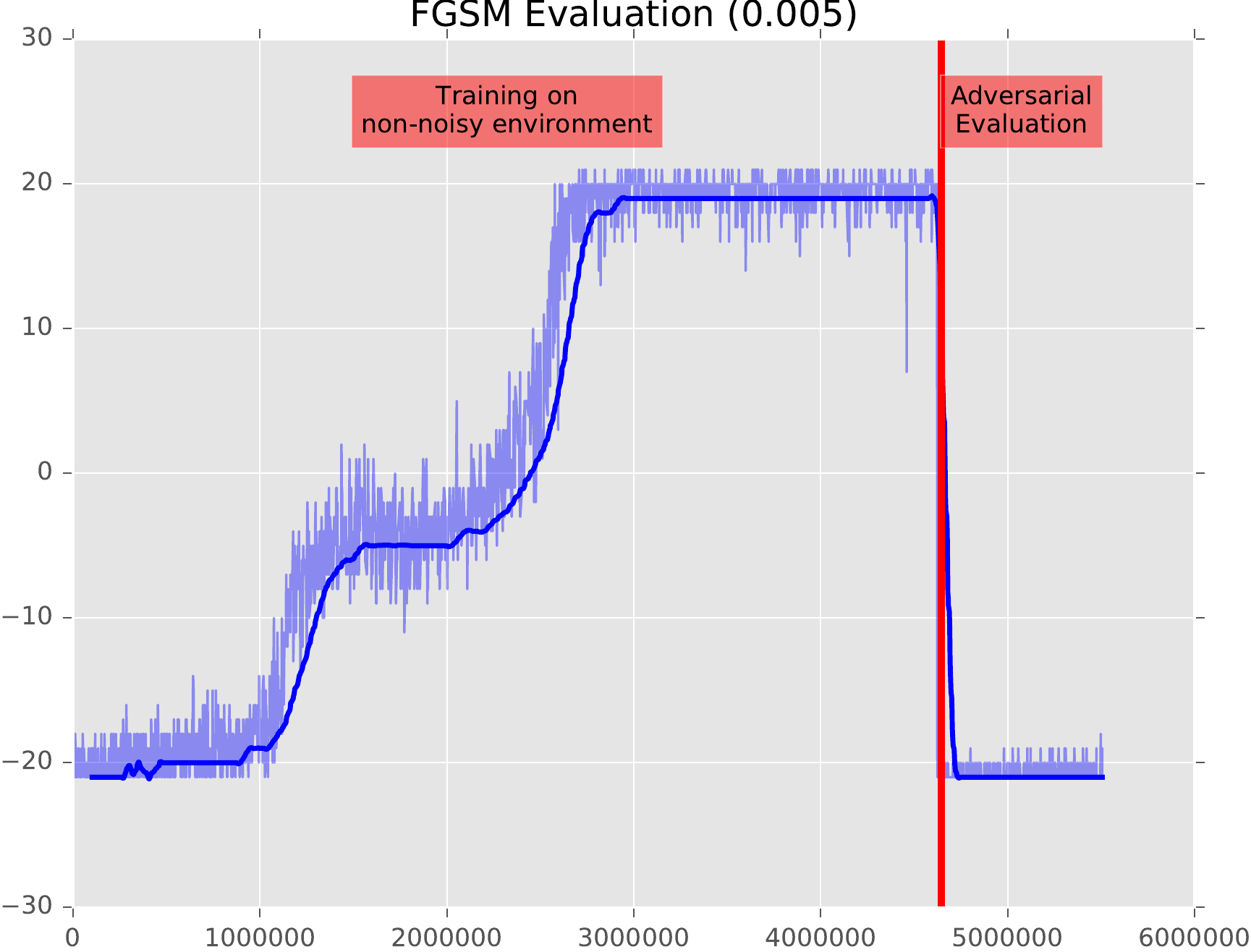}
\end{center}
\caption{Attack effectiveness of random noise with $\beta$ values $0.02$ and $0.05$ (top) vs. attack effectiveness of FGSM adversarial perturbations with $\epsilon$ values $0.001$ and $0.005$ (bottom).}
\label{fig:fixed-random-vs-adversarial}
\end{figure}

\begin{figure}[h]
\begin{center}
\includegraphics[scale=0.26]{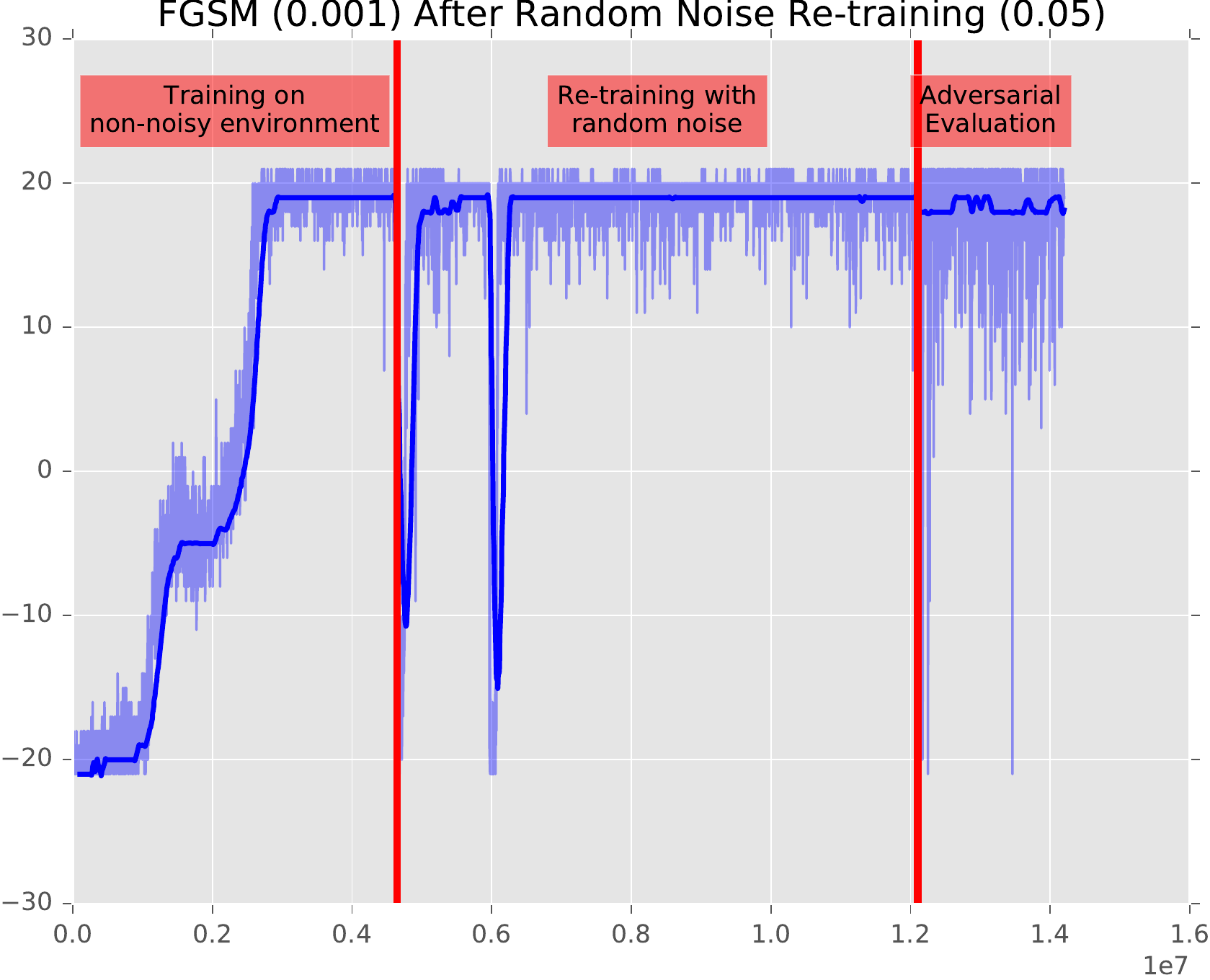}
\includegraphics[scale=0.26]{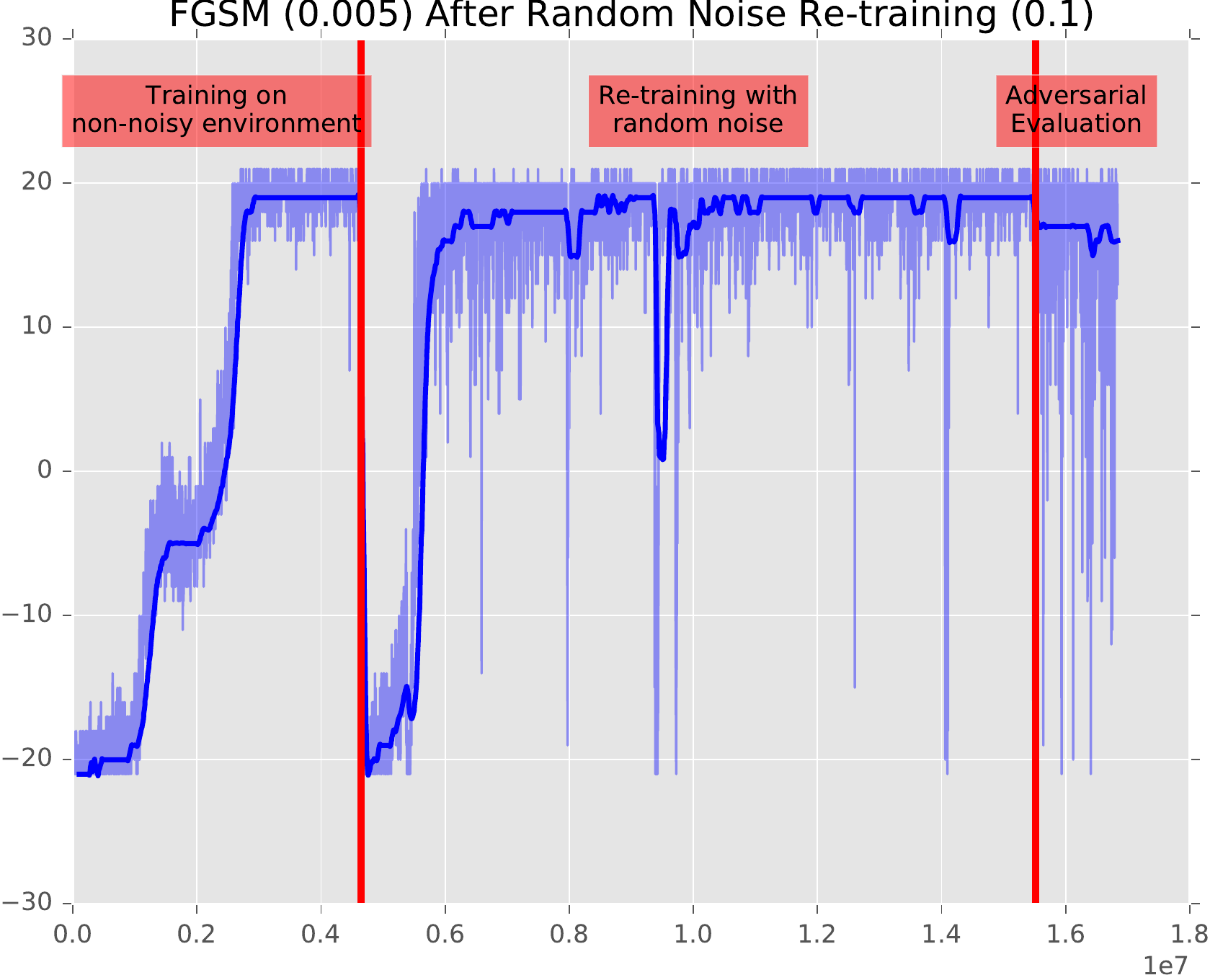}
\includegraphics[scale=0.26]{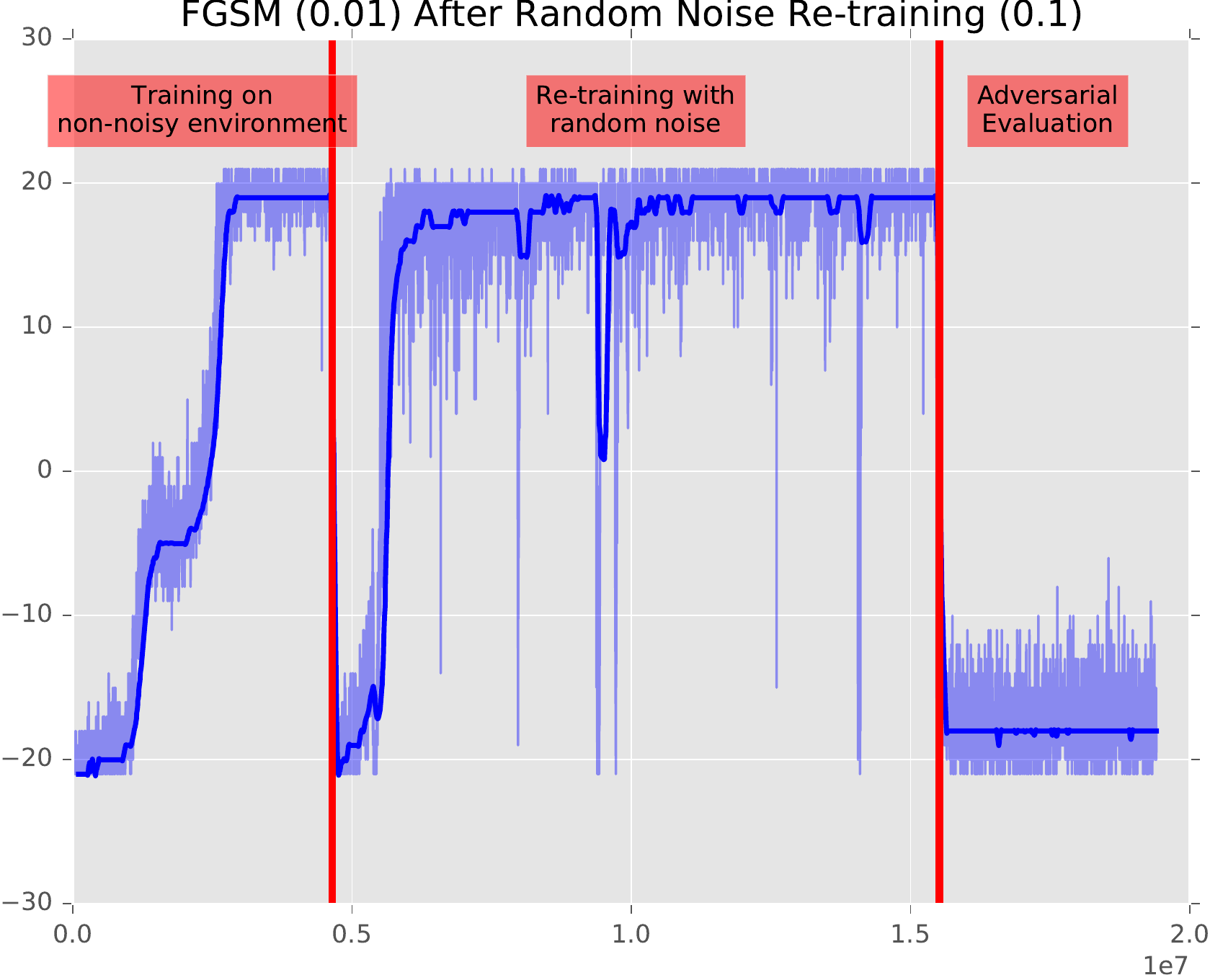} \\
\includegraphics[scale=0.26]{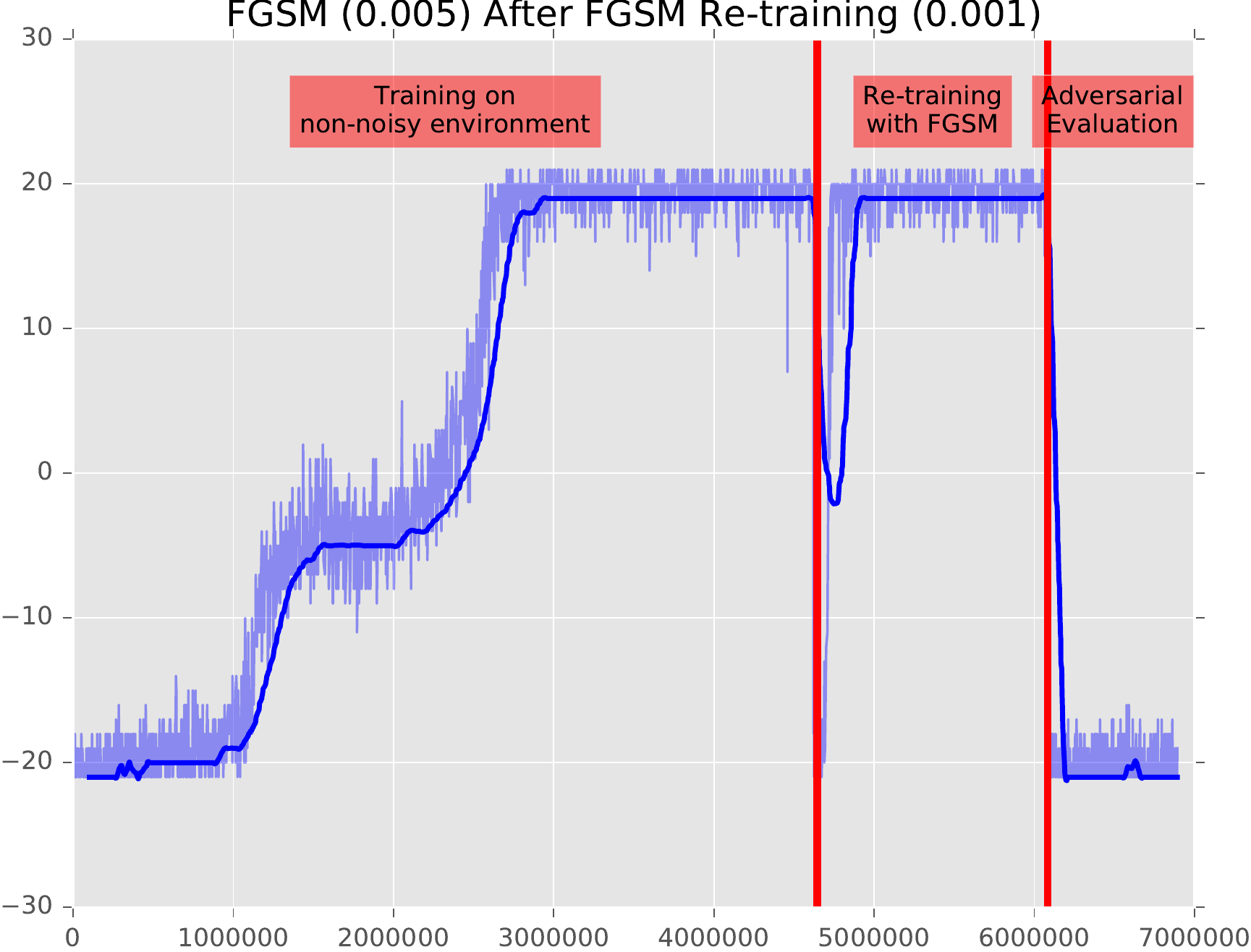}
\includegraphics[scale=0.26]{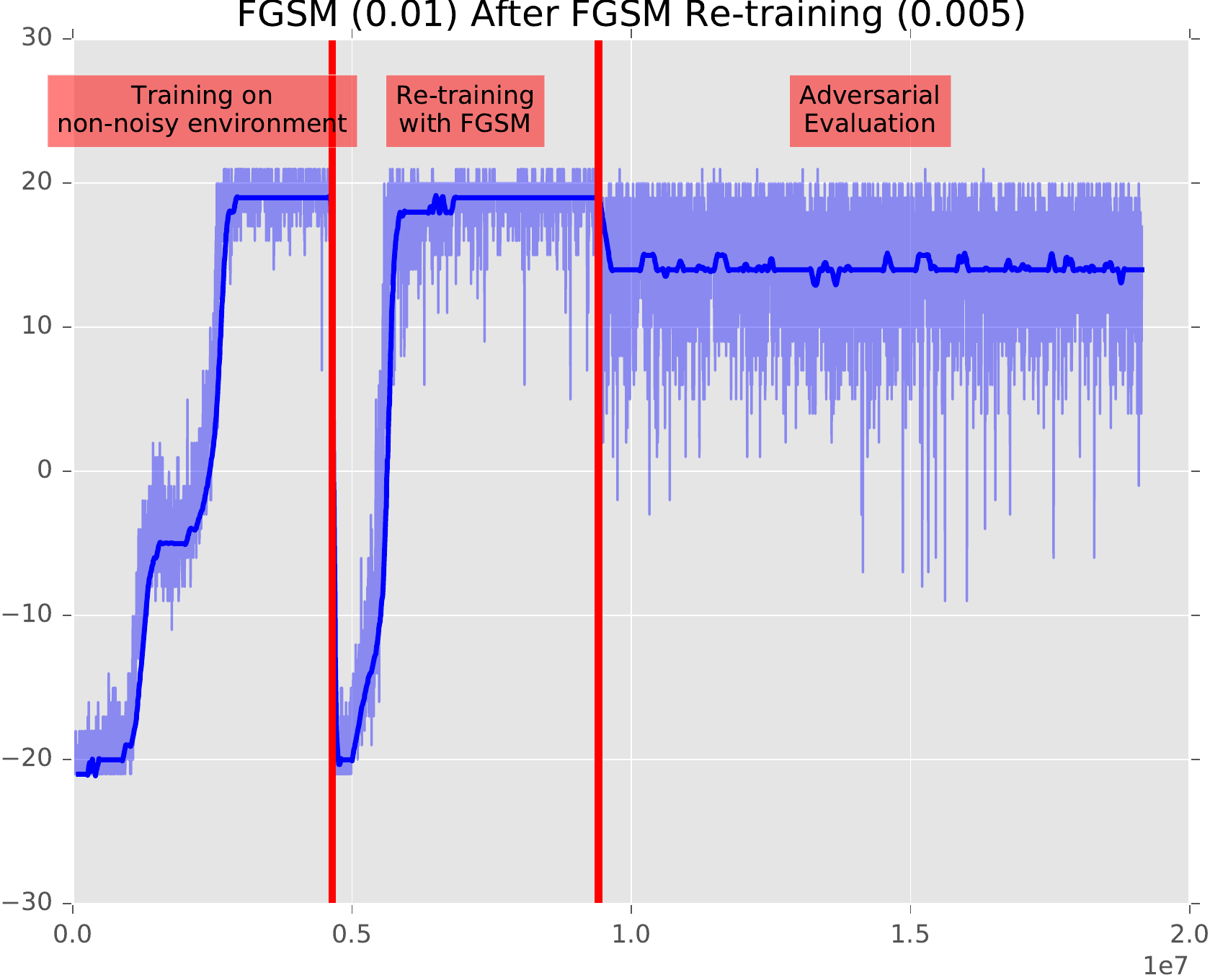}
\includegraphics[scale=0.26]{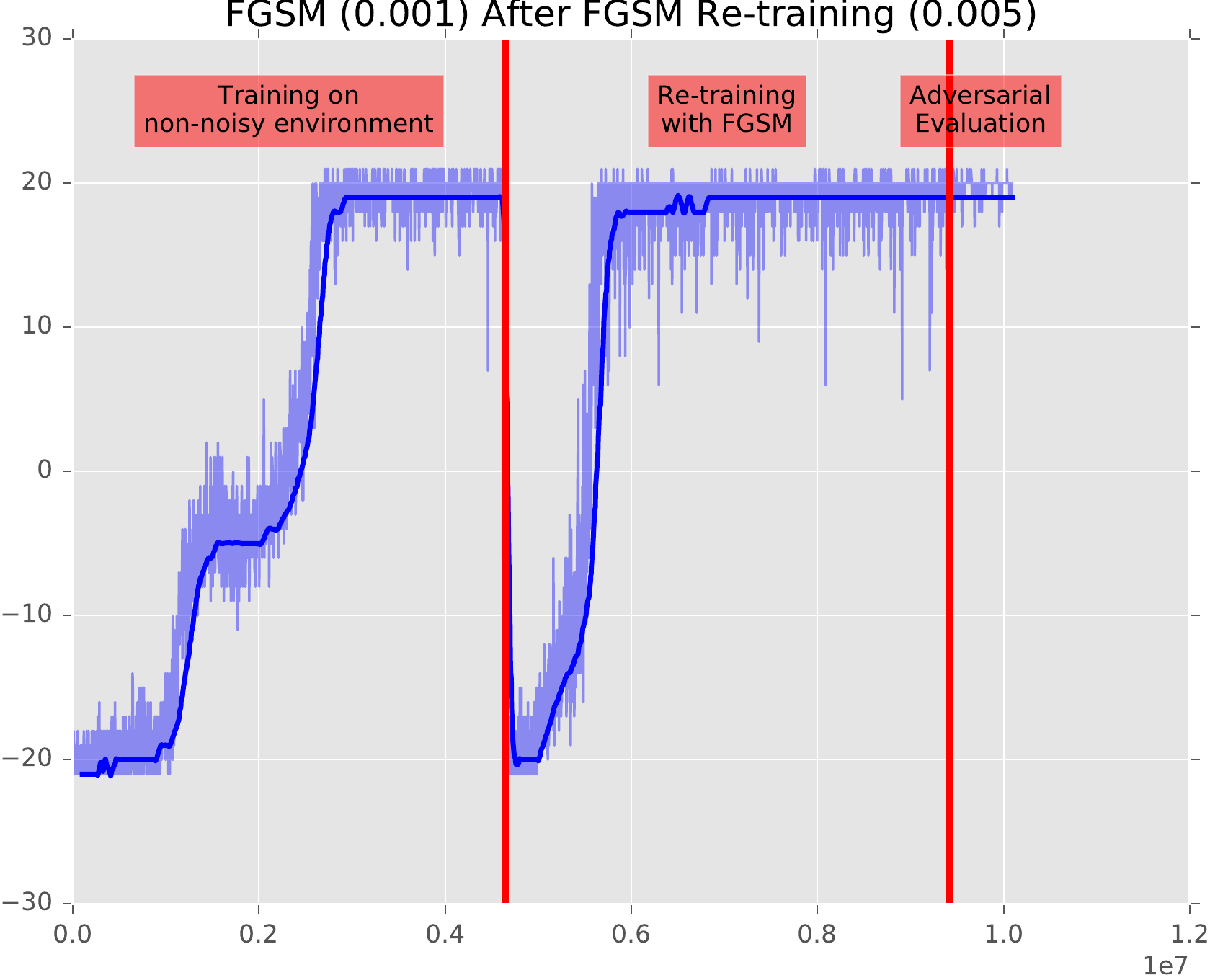}
\end{center}
\caption{Agent re-training experiments.
Initially the agent was trained on a non-noisy environment.
\textit{Top:} After first re-training with random noise (with $\beta$ values $0.05$ and $0.1$).
\textit{Bottom:} After first re-training with FGSM perturbations (with $\epsilon$ values $0.001$ and $0.005$).
After re-training, the agent was evaluated on FGSM perturbations (with $\epsilon$ values $0.001$, $0.005$ and $0.01$).
}
\label{fig:training-random-vs-adversarial}
\end{figure}

\subsection{Visualizing the policy network action boundary}

\begin{figure}[h]
\begin{center}
\raisebox{-0.5\height}{\includegraphics[scale=0.4]{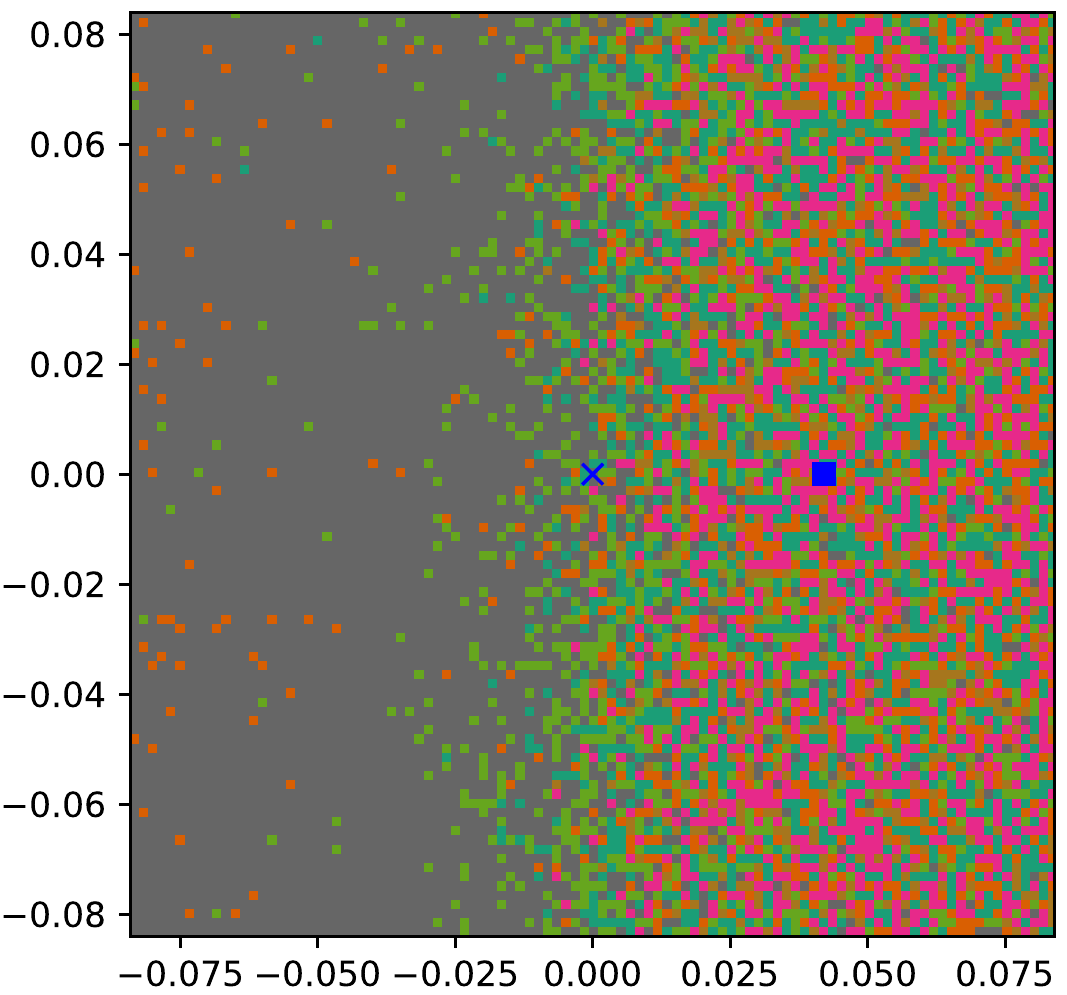}
\includegraphics[scale=0.4]{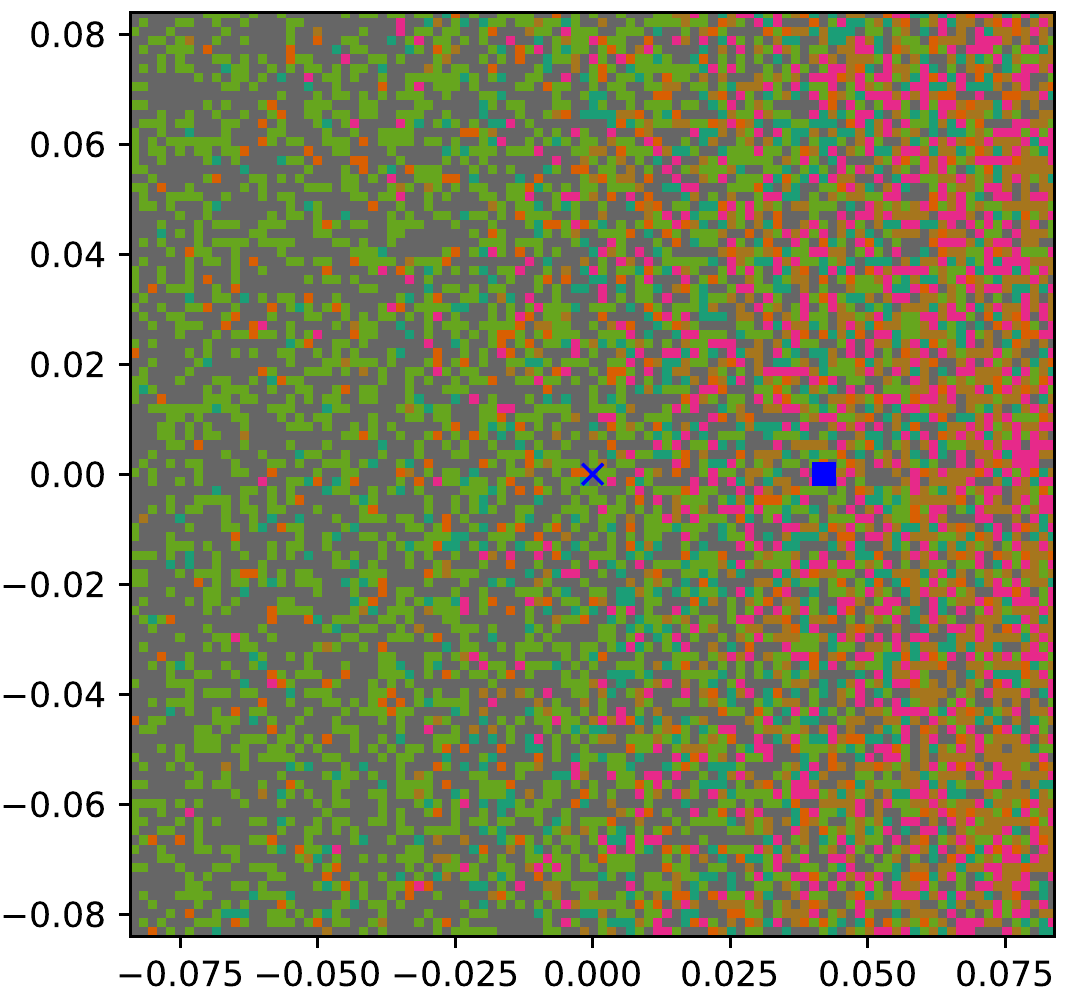}
\includegraphics[scale=0.4]{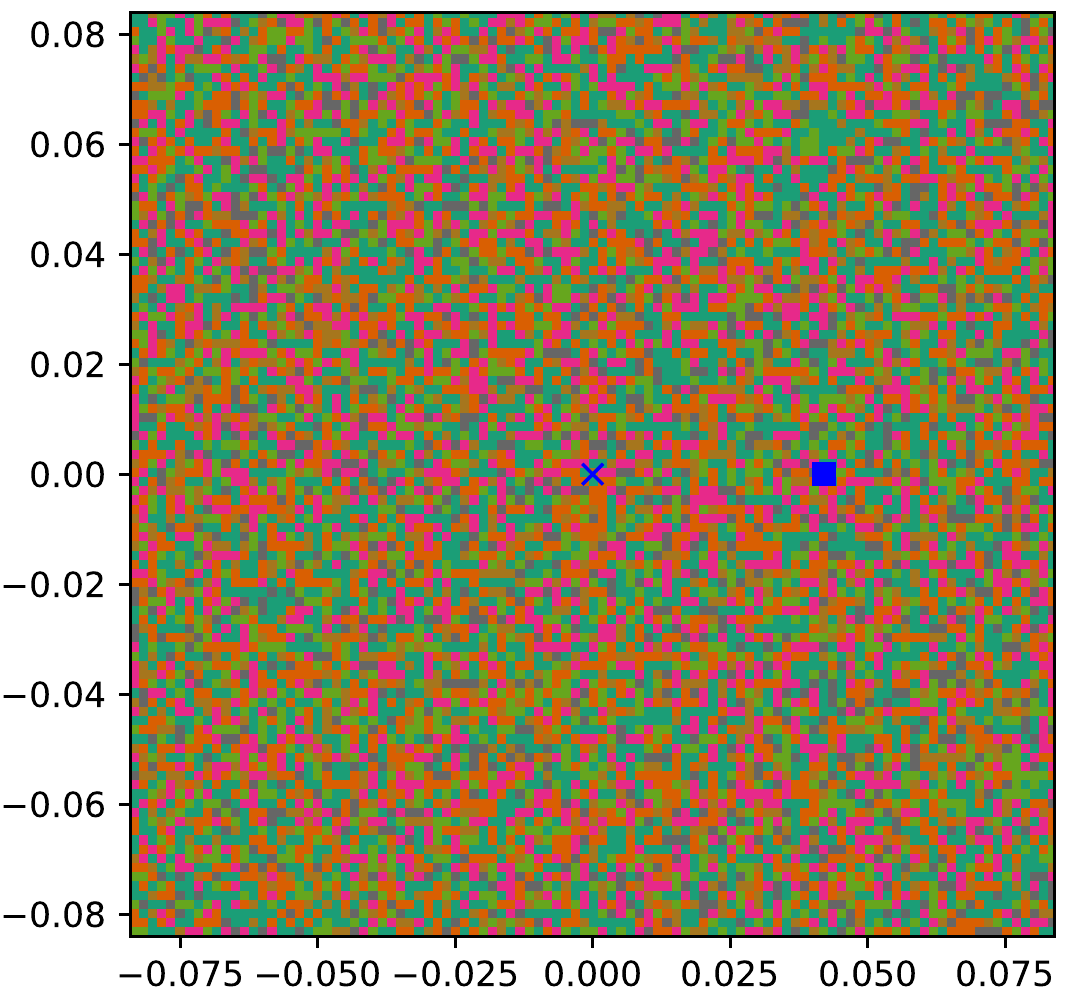}}
~
\raisebox{-0.5\height}{\includegraphics[scale=0.32]{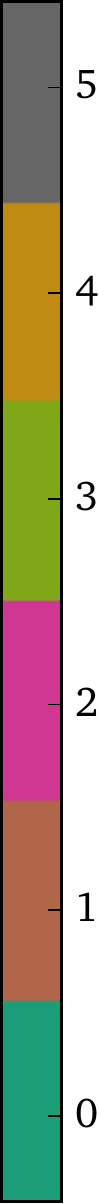}}
\end{center}
\caption{Visualization of actions in image space for a single frame.
The x-axis is in the direction of the generated adversarial example (FGSM, $\epsilon = 0.001$) for the target network.
The y-axis is in a random orthogonal direction.
Each point is the result of sampling the policy network $7$ times given the image at that point as input,
and shows the action most commonly output by the network (different actions have different colors).
Blue ``x'' marks the position of the original frame, while the blue square marks the position of the adversarial example.
Color bar on the far right shows the mapping of colors to discrete actions.
\textit{Left:} Baseline network without any re-training (action for the original input is action $5$).
\textit{Middle:} Network with re-training on random noise ($\beta = 0.1$, action for the original input is action $5$).
\textit{Right:} Network with re-training on FGSM perturbations ($\epsilon = 0.005$, action for the original input is action $0$).
}
\label{fig:action-boundary}
\end{figure}

We further study how the action boundary looks like for the policy network and how re-training affects it.
To this end, we prepare a visualization of predicted actions in image space (Figure~\ref{fig:action-boundary}).
We generate the plot by defining two normalized vectors, $\vd_1$ and $\vd_2$, spanning the input image space.
The one shown on the x-axis points in the direction of the generated adversarial perturbation ($\vd_1$), while the other shown on the y-axis points in a randomly chosen orthogonal direction ($\vd_2$).
The points in the plane represent actions predicted by the policy network for input $\vx + u \vd_1 + v \vd_2$, where $\vx$ is the original image (a single frame).
Since A3C is stochastic, we sample the predictions from the policy network $7$ times for each input and show the most common action.
Each discrete action (action $0$ to $5$) is represented by its own color, shown in the figure on the far right.
Values on the axes are the values of variables $u$ and $v$.

The visualization shows that the decision space is fragmented.
Small perturbations in the input can cause the optimal action chosen by the policy network to change drastically.
Re-training under a noisy environment (both random noise and adversarial FGSM perturbations) does not seem to make the decision boundary more smooth and the space seems to become even more fragmented (Figure~\ref{fig:action-boundary} middle and right).

\begin{figure}[h]
\begin{center}
\includegraphics[scale=0.4]{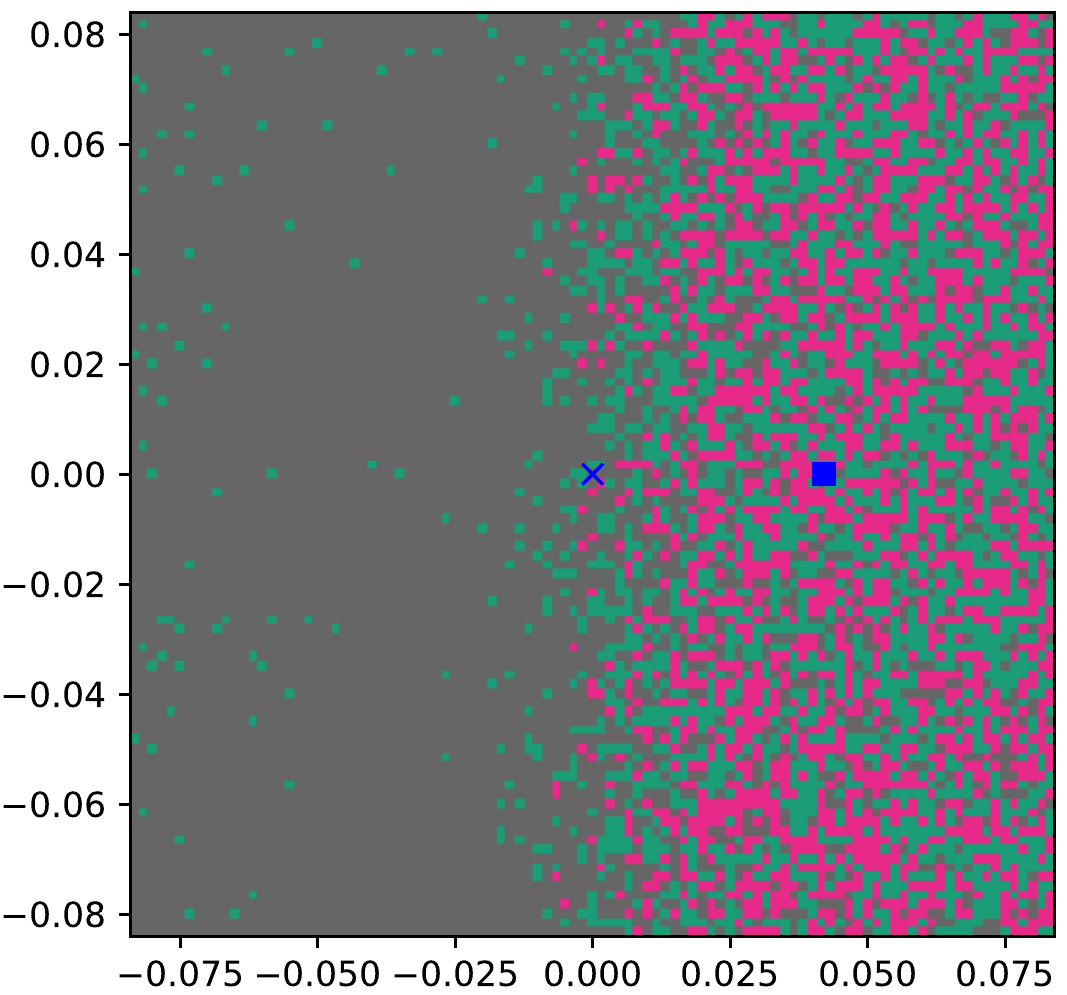}
\includegraphics[scale=0.4]{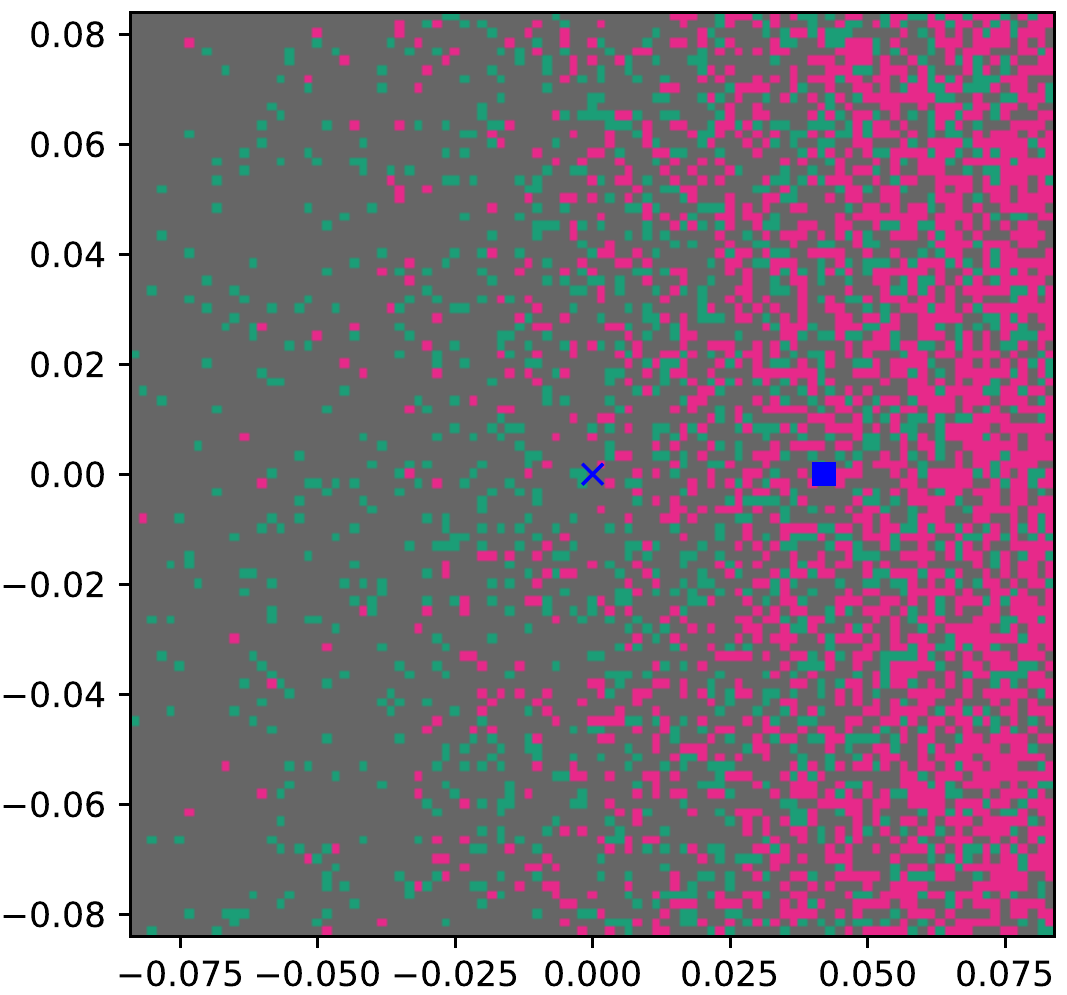}
\includegraphics[scale=0.4]{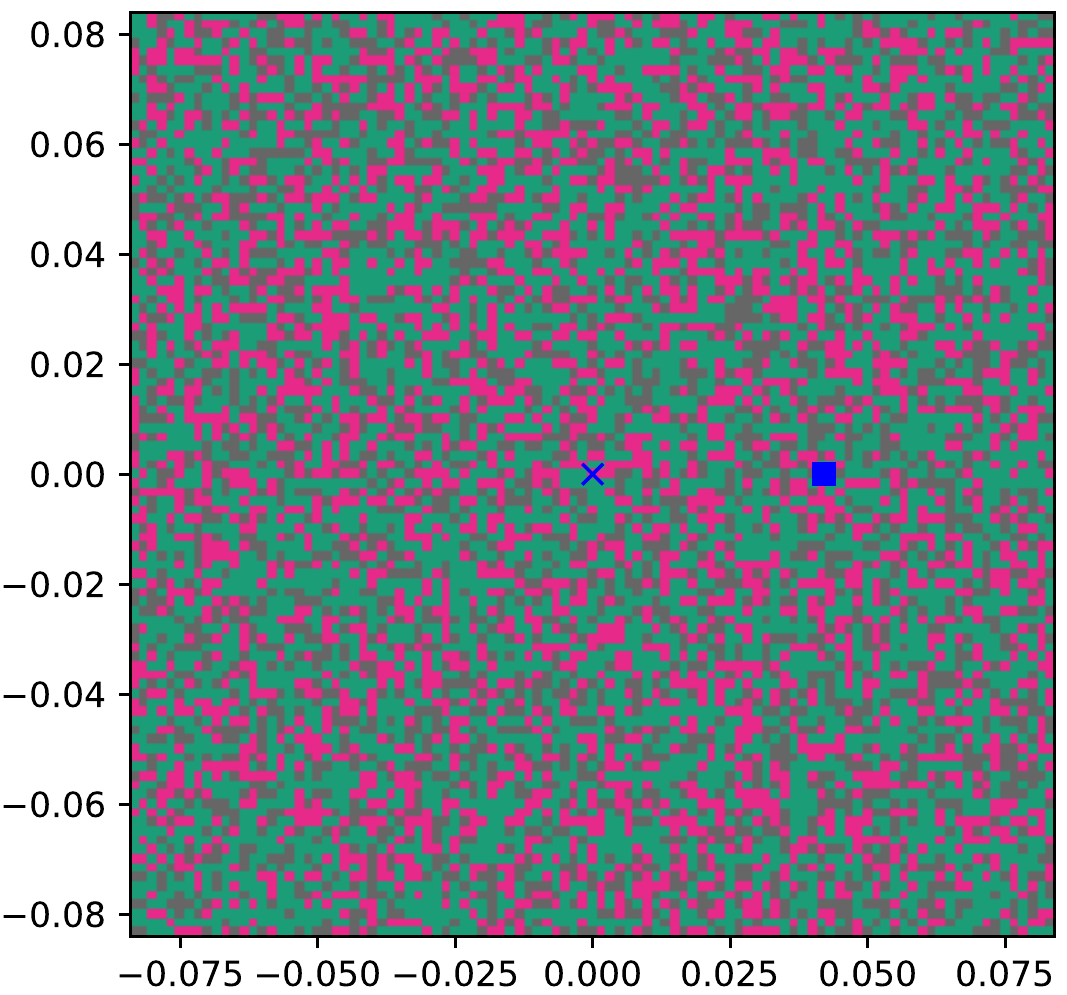} \\
\vspace{0.2cm}
\includegraphics[scale=0.5]{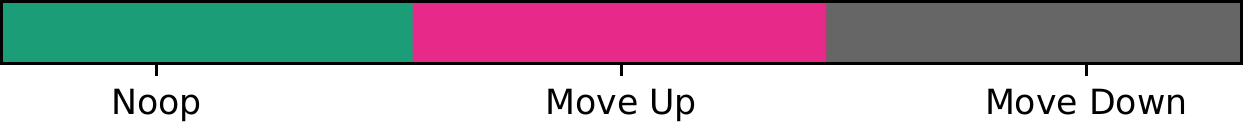}
\end{center}
\caption{Visualization of actions, adjusted for action semantics, in image space for a single frame.
The x-axis is in the direction of the generated adversarial example (FGSM, $\epsilon = 0.001$) for the target network.
The y-axis is in a random orthogonal direction.
Each point is the result of sampling the policy network $7$ times given the image at that point as input,
and shows the action most commonly output by the network (different actions have different colors, mapped based on action semantics).
Blue ``x'' marks the position of the original frame, while the blue square marks the position of the adversarial example.
Color bar below shows the mapping of colors to discrete actions.
\textit{Left:} Baseline network without any re-training (action for the original input is ``move down").
\textit{Middle:} Network with re-training on random noise ($\beta = 0.1$, action for the original input is ``move down").
\textit{Right:} Network with re-training on FGSM perturbations ($\epsilon = 0.005$, action for the original input is ``noop").
}
\label{fig:action-boundary-adjusted}
\end{figure}

We also adjust the visualization for action semantics.
The reasoning behind this is that even though the action space contains $6$ valid actions, the actions are actually duplicated (e.g., multiple actions actually have the exact same effect on the environment).
We manually checked the effect of each action on the environment and mapped the actions accordingly.
The three actions are: noop (do nothing), move the paddle up and move the paddle down.
Figure~\ref{fig:action-boundary-adjusted} shows that even when we adjust for duplication, the space remains fragmented.

\end{document}